\documentclass[10pt,twocolumn,letterpaper]{article}

\usepackage{wacv}
\usepackage{graphicx}
\usepackage{amsmath}
\usepackage{amssymb}
\usepackage{booktabs}
\usepackage{xcolor}
\usepackage[pagebackref,breaklinks,colorlinks]{hyperref}
\usepackage[capitalize]{cleveref}

\usepackage{array}
\usepackage{tabularx}
\newcolumntype{Y}{>{\raggedright\arraybackslash}X}
\crefname{section}{Sec.}{Secs.}
\Crefname{section}{Section}{Sections}
\Crefname{table}{Table}{Tables}
\crefname{table}{Tab.}{Tabs.}

\def\wacvPaperID{*****}
\def\confName{WACV}
\def\confYear{2025}

\title{\textbf{SGS: Segmentation-Guided Scoring for Global Scene Inconsistencies}}

\author{Gagandeep Singh\textsuperscript{1}, Samudi Amarsinghe\textsuperscript{1}, Urawee Thani\textsuperscript{1}, Ki Fung Wong\textsuperscript{1},\\
Priyanka Singh\textsuperscript{1}, Xue Li\textsuperscript{1}
}

\begin{document}
\maketitle

%%%%%%%%% ABSTRACT
\begin{abstract}
We extend HAMMER, a state-of-the-art model for multimodal manipulation detection, to handle global scene inconsistencies such as foreground--background (FG--BG) mismatch. While HAMMER achieves strong performance on the DGM\textsuperscript{4} dataset, it consistently fails when the main subject is contextually misplaced into an implausible background. We diagnose this limitation as a combination of label-space bias, local attention focus, and spurious text–foreground alignment. To remedy this without retraining, we propose a lightweight segmentation-guided scoring (SGS) pipeline. SGS uses person/face segmentation masks to separate foreground and background regions, extracts embeddings with a joint vision–language model, and computes region-aware coherence scores. These scores are fused with HAMMER’s original prediction to improve binary detection, grounding, and token-level explanations. SGS is inference-only, incurs negligible computational overhead, and significantly enhances robustness to global manipulations. This work demonstrates the importance of region-aware reasoning in multimodal disinformation detection. We release scripts for segmentaion and scoring at \url{https://github.com/Gaganx0/HAMMER-sgs}
\end{abstract}

%%%%%%%%% 1. INTRODUCTION
\section{Introduction}
Online disinformation increasingly exploits multimodal media---news-style images paired with persuasive captions---to manipulate public opinion. If applied politically, it could be used in information warfare campaigns and threaten democracy \cite{freelon2020disinfo}. It also spreads 6 times faster than true news \cite{vosoughi2018spread}. The combination of text and visuals yields surface plausibility: the content appears coherent at first glance, even when the underlying semantics are false. Such multimodal falsehoods are difficult to detect because they leverage mutually reinforcing modalities.

Recent benchmarks like DGM\textsuperscript{4} \cite{shao2023dgm4} stimulated progress in multimodal forgery detection by introducing face- and text-centric manipulations with grounding labels. HAMMER \cite{shao2023dgm4}, a transformer-based model with hierarchical reasoning, set a strong baseline by jointly detecting manipulations, classifying types, and grounding altered regions or tokens. Yet HAMMER and DGM\textsuperscript{4} share a critical blind spot: they lack coverage of \emph{global} manipulations such as mismatched foregrounds and backgrounds. 

Consider a politician giving a speech on the surface of Mars or a teacher calmly instructing students inside a collapsing glacier cave. These examples, while visually plausible, exhibit glaring contextual implausibility. Current detectors, trained on local manipulations (face swaps, attributes, caption edits), miss such cases. Our motivation is to explicitly address this gap.

\paragraph{Contributions.}
\begin{itemize}
    \item We diagnose HAMMER’s systematic failures on FG--BG mismatches, identifying causes in its label space, attention distribution, and cross-modal biases.
    \item We propose segmentation-guided scoring (SGS), a lightweight, inference-only extension that introduces region-aware consistency signals without retraining.
    \item SGS combines foreground–text, background–text, and foreground–background scores with HAMMER’s native output, improving detection and grounding.
    \item We provide an evaluation protocol on DGM\textsuperscript{4} and our FG--BG extension.
\end{itemize}

%%%%%%%%% 2. RELATED WORK
\section{Related Work}
\textbf{Multimodal manipulation benchmarks.}  
Early research into detecting manipulated content focused primarily on unimodal datasets. For example, FaceForensics++ \cite{rossler2019faceforensicspp}
was fundamental for analyzing facial forgeries, while LIAR \cite{wang2017liar} provided a benchmark for detecting false information in short political statements. These initial efforts, while valuable, were limited in scope.

As the research field progressed, the focus shifted toward multimodal datasets that integrated multiple modalities to better simulate realistic manipulations. NewsCLIPpings \cite{luo2021newsclippings} and COSMOS \cite{aneja2023cosmos} were innovative works that mimic out-of-context news pairs which challenged models to identify inconsistencies between images and their accompanying text. AV-Deepfake++ \cite{cai2025avdeepfakepp} and MFND \cite{zhu2025mfnd} on the other hand introduced audiovisual components to enable the detection of deepfakes that manipulate both video and sound, in addition to expanded fake news detection scenarios.

DGM\textsuperscript{4} remains the most comprehensive multimodal benchmark, modeling a variety of local manipulations. However, all existing benchmarks fail to systematically introduce FG-BG mismatches.

\textbf{Detection and grounding methods.} 
Approaches for detecting and grounding manipulations can be broadly categorized into several techniques. The first method is self-supervised alignment \cite{aneja2023cosmos}, where models learn to align textual captions with specific objects within an image without explicit labels provided. This enables the detection of subtle inconsistencies by examining the inherent relationships between different data types.

The second technique is semantic consistency learning \cite{zhu2025mfnd,zhang2024asap}, which trains a model to spot and penalize semantic discrepancies between an image and the meaning of its accompanying text. A semantic mismatch strongly indicates manipulation.

Frequency-aware modeling \cite{liu2025ufa} relies on a different principle by examining the spatial structure of images by decomposing images into frequency sub-bands, effectively detecting manipulation and forgery artifacts.

HAMMER notably contributes by integrating hierarchical reasoning with grounding to provide fine-grained localization of manipulated regions. However, it is constrained by its training distribution, making it less effective on novel manipulation types. Our method complements HAMMER with an external consistency signal to enhance detection accuracy and robustness on examples with atypical distributions.

\textbf{Region-level reasoning.}  
Region-level reasoning leverages segmentation to separate salient objects from backgrounds \cite{kirillov2023sam}. In multimodal detection, this enables localized reasoning. By using segmentation, a model can independently compare foreground and background features against text. We adopt this principle by using SAM \cite{kirillov2023sam} (Segment Anything Model) to generate precise masks that isolate objects, then feed into CLIP \cite{radford2021clip} to compute and compare embeddings, allowing us to locate contextual inconsistencies between the foreground, background and text.

%%%%%%%%% 3. WHY HAMMER FAILS
\section{Why HAMMER Fails on FG--BG Mismatch}

HAMMER is trained and evaluated on four manipulation types in DGM\textsuperscript{4} (FS/FA/TS/TA) \cite{shao2023dgm4}, all of which are predominantly \emph{local} edits. Foreground--background (FG--BG) mismatch---the subject is plausible but the surrounding scene is semantically out of place---is not represented in the label space, yielding a systematic blind spot. In practice, cross-modal attention focuses on faces and caption tokens; when faces are photorealistic and explicitly referenced, the background receives less influence on the binary decision.

To probe this failure mode \emph{without retraining or modifying HAMMER}, we use a lightweight, stand-alone semantic check that inspects only the \emph{relationship between foreground and background}. The method relies exclusively on two inputs per sample: a foreground crop and a background crop. No additional textual supervision is required.

\begin{figure}[t]
    \centering
    \includegraphics[width=\linewidth]{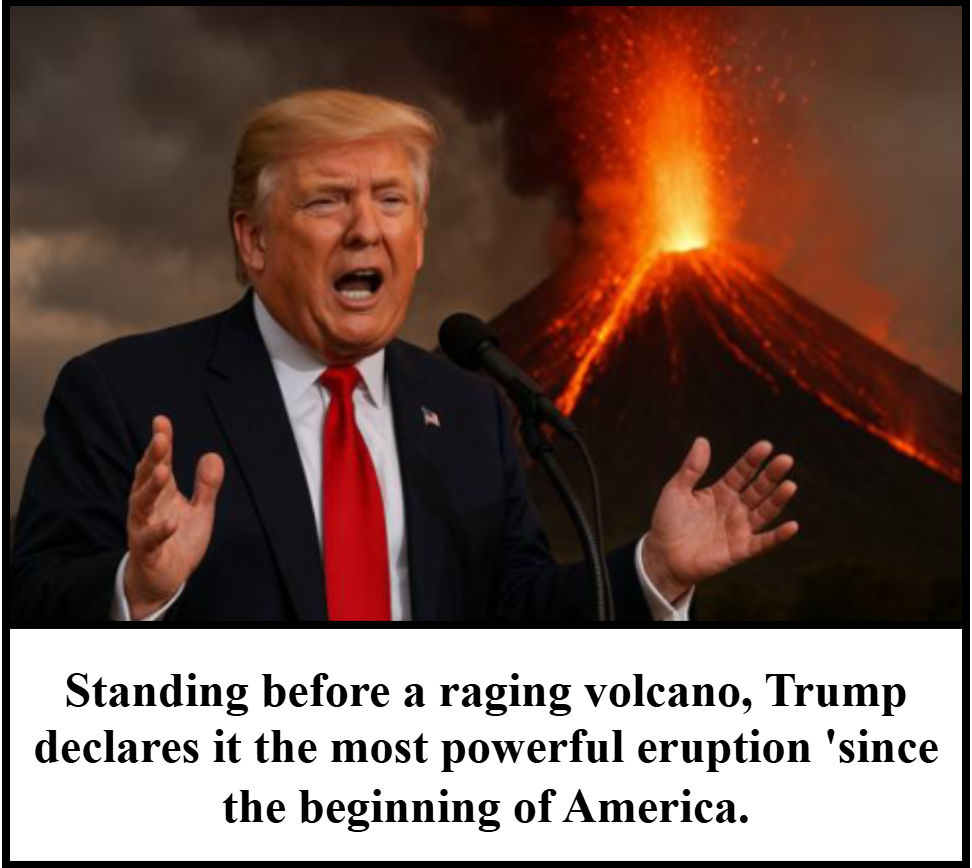}
    \caption{Illustrative failure: subject and caption align locally, but the scene is globally implausible. Our checker targets the FG--BG relationship directly.}
    \label{fig:fail-example}
\end{figure}

%%%%%%%%% 4. METHODOLOGY
\section{Methodology}

\begin{figure*}[t]
    \centering
    \includegraphics[width=\linewidth]{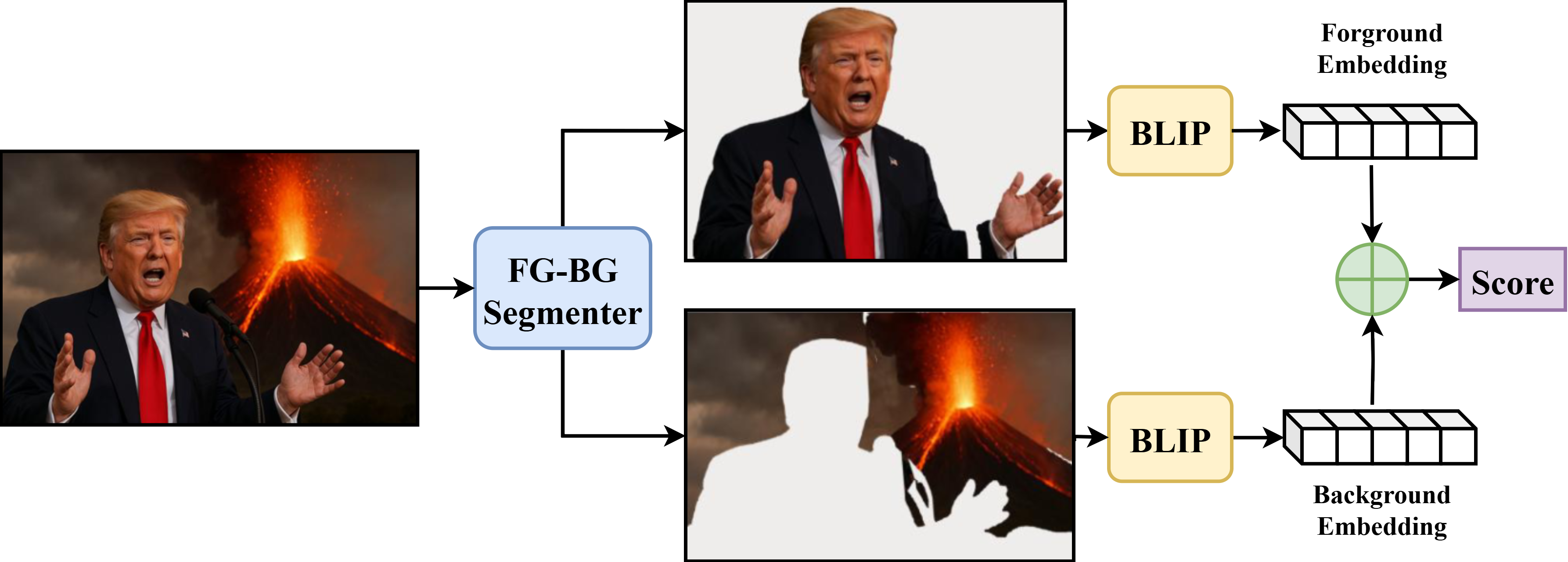}
    \caption{Segmentation-Guided Scoring (SGS) pipeline. Any segmenter can supply the foreground and background crops. Each crop is captioned independently (BLIP), embedded into a semantic space (MiniLM), and compared via cosine similarity. A low similarity score signals FG--BG inconsistency.}
    \label{fig:sgs-pipeline}
\end{figure*}

Our goal is to design a lightweight, inference-only module that detects global foreground--background (FG--BG) inconsistencies without retraining. The proposed Segmentation-Guided Scoring (SGS) pipeline (Fig.~\ref{fig:sgs-pipeline}) relies on region decomposition, captioning, and text-space similarity. Unlike prior vision-only metrics, SGS translates both subject and scene into natural language descriptions, enabling us to directly compare their semantics using a compact sentence encoder. Below we describe the components in detail.

\paragraph{Inputs.} 
Each input image $I$ is assumed to have two precomputed crops: a \emph{foreground} crop $I^{\text{FG}}$ representing the detected person(s), and a \emph{background} crop $I^{\text{BG}}$ capturing the surrounding context. These crops may be obtained by any segmentation or detection model (e.g., SAM, YOLOv8-seg), or provided externally. Our released tool does not perform segmentation itself; it simply consumes the paths to these crops and treats them as the atomic units for reasoning.

\paragraph{Captioning the two views.}
For both $I^{\text{FG}}$ and $I^{\text{BG}}$, we generate short, neutral textual descriptions using BLIP (base) in image-captioning mode:
\[
c^{\text{FG}} = \mathrm{BLIP}(I^{\text{FG}}), \qquad
c^{\text{BG}} = \mathrm{BLIP}(I^{\text{BG}}).
\]
BLIP operates with implicit prompts; no handcrafted instructions are used, ensuring reproducibility and avoiding prompt bias. To standardize computation, all crops are resized to $448{\times}448$ before feeding into BLIP, which then applies its own internal resizing. This step produces captions that summarize “who/what” is present in the foreground and “where” they appear in the background.

\paragraph{Semantic similarity in text space.}
We then embed both captions into a common semantic space using MiniLM (all-MiniLM-L6-v2), a lightweight sentence transformer. Let $\phi(\cdot)$ denote the encoder. The cosine similarity
\[
s_{FB} \;=\; \cos\!\Big(\phi(c^{\text{FG}}), \phi(c^{\text{BG}})\Big)
\]
measures how semantically aligned the subject and context are. Because cosine values may range in $[-1,1]$, we apply a simple post-processing step to map scores into $[0,1]$:
\[
\tilde{s}_{FB} \;=\; 
\begin{cases}
\min(1, s_{FB}), & s_{FB} \ge 0, \\[4pt]
\max\!\big(0, \tfrac{s_{FB}+1}{2}\big), & s_{FB} < 0.
\end{cases}
\]
This normalized score $\tilde{s}_{FB}$ is stored as \texttt{sts01} in the output CSV. High values indicate agreement between FG and BG, while low values suggest implausibility.

\paragraph{Decision rule.}
We adopt a one-class formulation: all test samples are evaluated against a fixed threshold $\tau$ (default $\tau{=}0.55$, configurable at runtime). The checker outputs a binary label
\[
\texttt{label} =
\begin{cases}
\texttt{Match}, & \tilde{s}_{FB} \ge \tau, \\
\texttt{Mismatch}, & \tilde{s}_{FB} < \tau.
\end{cases}
\]
A mismatch implies that the subject’s semantics are poorly supported by the background description, which corresponds to global inconsistency. The output row is written as:
\{\texttt{id, fg\_path, bg\_path, fg\_text, bg\_text, sts01, label}\}.

\begin{figure}[t]
    \centering
    \includegraphics[width=\linewidth]{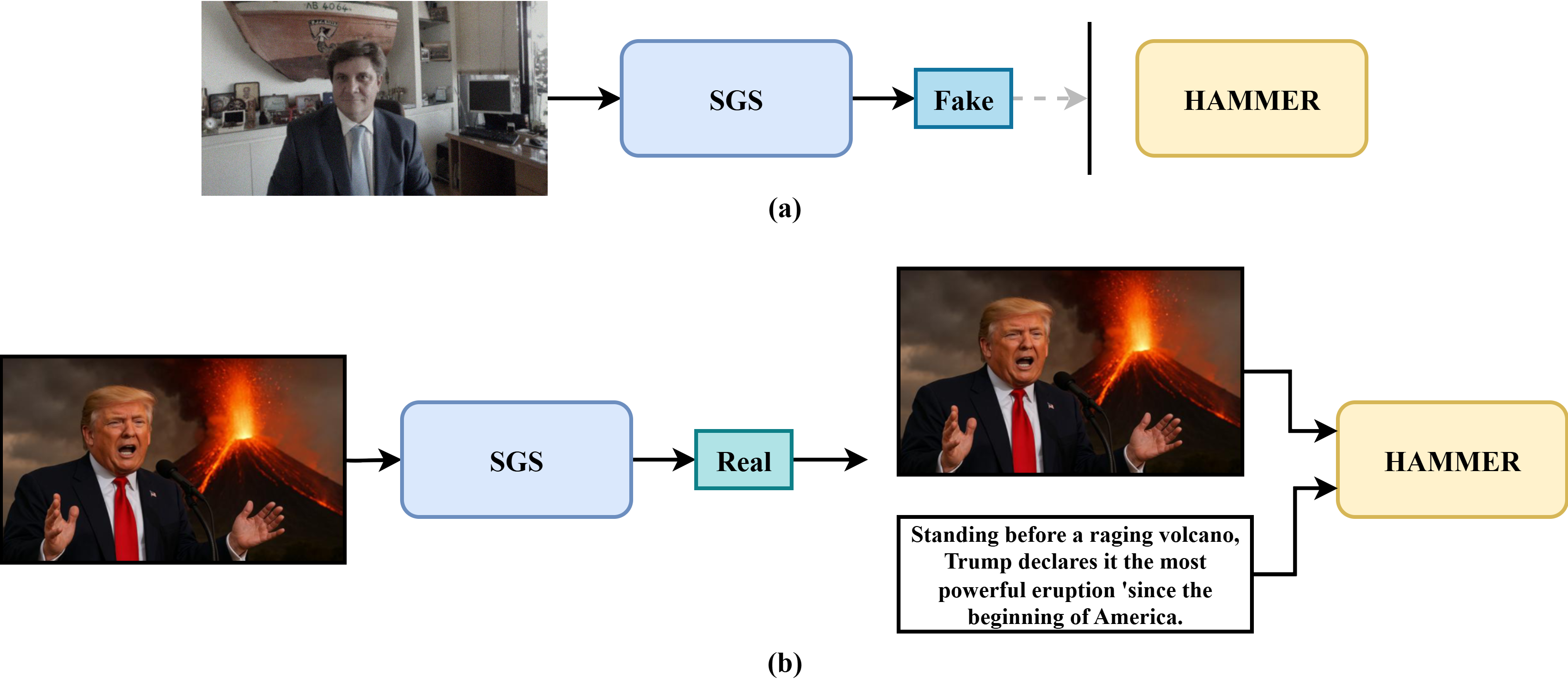}
    \caption{Integration of SGS with HAMMER. (a) If SGS deems the subject–scene pair consistent, HAMMER is applied directly. (b) If SGS flags a mismatch, the image and caption are not routed into HAMMER, saving computational power.}
    \label{fig:sgs-cases}
\end{figure}

\paragraph{Integration with HAMMER.}
While SGS provides a coarse plausibility filter, it is complementary to HAMMER’s fine-grained manipulation reasoning. We integrate SGS in front of HAMMER as shown in Fig.~\ref{fig:sgs-cases}. If SGS predicts a \emph{Match}, the input is passed directly to HAMMER for local manipulation detection. If SGS predicts a \emph{Mismatch}, the image and its caption are explicitly flagged as suspicious and routed through HAMMER for deeper grounding and token-level analysis. This cascaded design combines the efficiency of a lightweight text-similarity check with the interpretability of a transformer-based multimodal detector.

\paragraph{Implementation details.}
Our prototype uses \texttt{Salesforce/blip-image-captioning-base} for caption generation and \texttt{sentence-transformers/all-MiniLM-L6-v2} for similarity scoring, both running in \texttt{fp32} mode. Crops are loaded with PIL, resized to $448{\times}448$, captioned independently, and encoded for cosine similarity. The checker supports three pairing modes: (i) a CSV file with \texttt{id, fg, bg} columns, (ii) a JSON list of identifiers with automatic extension resolution, or (iii) auto-pairing by filename stems across \texttt{fg\_dir} and \texttt{bg\_dir}. All configuration flags, defaults, and output headers mirror the released script exactly, ensuring reproducibility.

%%%%%%%%% 5. EXPERIMENTS
\section{Experiments}
\label{sec:experiments}

We study a minimal, modular pipeline that surfaces global FG--BG inconsistency without retraining any base detector. The pipeline has two pluggable stages: (1) a \emph{subject/context splitter} that yields a foreground crop (FG) and a complementary background crop (BG) per image (e.g., any detector/segmenter such as SAM or YOLO-seg, or precomputed crops); and (2) a \emph{semantic comparator} that converts FG and BG into short texts and measures their similarity. In all experiments reported here we \emph{instantiate} the comparator \textbf{exactly as in our released script}: BLIP-base for captioning each crop independently and MiniLM (all-MiniLM-L6-v2) for cosine similarity between the two captions, followed by a fixed threshold.

\subsection{Why these baselines?}
We compare to four complementary families that probe FG--BG mismatch from different angles:
\begin{enumerate}
\item \textbf{Contrastive VLM encoders} (OpenCLIP \cite{cherti2023openclip}, SigLIP \cite{zhai2023siglip}) test whether role text aligns more with FG than BG, operationalizing global compatibility as a simple similarity \emph{gap} without full reasoning.
\item \textbf{Vision-only self-supervised features} (DINOv2 \cite{oquab2024dinov2}) assess whether FG and BG lie far apart in feature space, isolating a purely visual signal that ignores captions.
\item \textbf{Multimodal LLM} (Qwen2-VL-2B-Instruct \cite{qwen2vl2024}) answers a strict Yes/No plausibility question, offering a reasoning-style signal.
\item \textbf{Task-specific detector} (HAMMER \cite{shao2023dgm4}) anchors results w.r.t.\ the widely used DGM\textsuperscript{4} model family trained on local manipulations (FS/FA/TS/TA).
\end{enumerate}

\subsection{Evaluation setup}
We evaluate on an Extended DGM\textsuperscript{4} \emph{inconsistent} split where \emph{all images are FG--BG mismatched}. Because this is an all-positive (one-class) setting, we report \emph{Acc} and \emph{F1} against all-1 labels and, where applicable, the fraction flagged as inconsistent. For our script, crops are provided externally; we run BLIP-base (\texttt{fp32}) with $16$ max tokens and MiniLM, exactly matching the public script defaults and I/O (\texttt{sts01} and \texttt{label}). Threshold $\tau$ is 0.55 unless otherwise stated.

\subsection{Our crop–caption checker (script)}
\begin{table}[!tb]
\centering
\caption{Our FG--BG crop–caption checker (BLIP-base + MiniLM), exactly as implemented in the released script. All images in the split are inconsistent.}
\label{tab:ours}
\vspace{-2mm}
\setlength{\tabcolsep}{4pt}\scriptsize
\begin{tabularx}{\linewidth}{@{}Ycccc@{}}
\toprule
\textbf{N} & $\tilde{s}_{FB}$ mean$\pm$sd & median & \% \texttt{Mismatch} & Acc / F1 \\
\midrule
4632 & $0.264\pm0.158$ & 0.238 & 94.3 & 0.943 / 0.971 \\
\bottomrule
\end{tabularx}
\end{table}

\paragraph{Reading Table~\ref{tab:ours}.}
$\tilde{s}_{FB}\in[0,1]$ is the normalized cosine similarity between the FG and BG captions produced by BLIP; lower indicates stronger semantic disagreement between subject and scene. The script’s binary label is \texttt{Mismatch} iff $\tilde{s}_{FB}<\tau$ with $\tau{=}0.55$.

\subsection{Baselines}
\paragraph{Contrastive vision--language encoders (gap test).}
We compute $\Delta=s(\mathrm{FG},\texttt{text})-s(\mathrm{BG},\texttt{text})$ using OpenCLIP and SigLIP with short role prompts and predict ``inconsistent'' if $\Delta<0$.

\begin{table}[!tb]
\centering
\caption{Vision--Language encoders on the all-inconsistent split (one-class). $\Delta = s(\text{FG},\text{text}) - s(\text{BG},\text{text})$.}
\label{tab:vl}
\vspace{-2mm}
\setlength{\tabcolsep}{4pt}\scriptsize
\begin{tabularx}{\linewidth}{@{}Yccc@{}}
\toprule
\textbf{Method} & $\Delta$ mean$\pm$sd & \% $\Delta<0$ & Acc / F1 \\
\midrule
OpenCLIP (ViT-B/32) & $-0.046\pm0.070$ & 74.3 & 0.743 / 0.852 \\
SigLIP (base/16)    & $-0.003\pm0.026$ & 53.9 & 0.539 / 0.701 \\
\bottomrule
\end{tabularx}
\end{table}

\paragraph{Vision-only distance (DINOv2).}
Foreground and background crops are embedded (ViT-B/14), and we measure $d=\lVert F-B\rVert_2$ after feature normalization. A one-class proxy threshold at the dataset median yields:

\begin{table}[!tb]
\centering
\caption{Vision-only baseline (DINOv2 ViT-B/14). FG–BG feature distance $d=\lVert F-B\rVert_2$. One-class proxy threshold at median.}
\label{tab:dino}
\vspace{-2mm}
\setlength{\tabcolsep}{4pt}\scriptsize
\begin{tabularx}{\linewidth}{@{}Ycc@{}}
\toprule
\textbf{Method} & Dist (mean$\pm$sd) & Acc / F1 @ median \\
\midrule
DINOv2 (ViT-B/14) & $1.244\pm0.030$ & 0.500 / 0.667 \\
\bottomrule
\end{tabularx}
\end{table}

\paragraph{Multimodal LLM (Yes/No).}
Qwen2-VL-2B-Instruct is prompted with a binary plausibility question; ``No'' is mapped to ``inconsistent'':

\begin{table}[!tb]
\centering
\caption{Qwen2-VL-2B-Instruct with Yes/No prompt (V2 framing).}
\label{tab:vlm}
\vspace{-2mm}
\setlength{\tabcolsep}{4pt}\scriptsize
\begin{tabularx}{\linewidth}{@{}Yccc@{}}
\toprule
\textbf{Method} & \% ``No'' & Acc & F1 \\
\midrule
Qwen2-VL-2B-Instruct & 32.8 & 0.328 & 0.494 \\
\bottomrule
\end{tabularx}
\end{table}

\paragraph{Task-specific anchor (HAMMER).}
We run the public HAMMER checkpoint; FG--BG is out-of-vocabulary for its type head.

\begin{table}[!tb]
\centering
\caption{HAMMER on FG--BG extension. FG--BG is out-of-vocabulary for its type head.}
\label{tab:hammer}
\vspace{-2mm}
\setlength{\tabcolsep}{3pt}\scriptsize
\begin{tabularx}{\linewidth}{@{}Ycccc@{}}
\toprule
\textbf{Method} & ACC$_{cls}$ & OF1 & IoU mean & Tok Acc / F1 \\
\midrule
HAMMER (released) & 19.1 & 35.3 & 94.9 & 78.8 / 22.0 \\
\bottomrule
\end{tabularx}
\end{table}

\subsection{Discussion}

\paragraph{Our crop–caption checker is a strong one-class probe.}
With no supervision beyond BLIP+MiniLM and no access to the news caption, the script flags $94.3\%$ of truly inconsistent cases (Table~\ref{tab:ours}). In an all-positive split this translates to $0.943$ accuracy and $0.971$ F1 with $\tau{=}0.55$. The low \(\tilde{s}_{FB}\) mean ($0.264$) indicates systematic semantic disagreement between subject and scene, aligning with our FG--BG hypothesis.

\paragraph{Contrastive encoders reveal a gap signal.}
OpenCLIP separates FG from BG via text alignment, with $74.3\%$ of samples yielding $\Delta{<}0$ (Table~\ref{tab:vl}). SigLIP is far weaker. This suggests the geometry of the learned joint space crucially affects sensitivity to global scene plausibility.

\paragraph{Vision-only distances capture stable separation.}
DINOv2 shows tight distributions of FG--BG distances (Table~\ref{tab:dino}), indicating a consistent visual notion of ``subject vs.\ context.'' While the naive median threshold gives balanced accuracy, the result implies that visual-only FG/BG statistics are useful complements to caption-based signals.

\paragraph{General-purpose VLMs under-report inconsistency.}
Qwen2-VL-2B-Instruct returns ``No'' only $32.8\%$ of the time on an all-inconsistent set (Table~\ref{tab:vlm}). This optimism bias is consistent with instruction-tuning for plausibility rather than adversarial detection.

\paragraph{Implications for a HAMMER extension.}
HAMMER localizes people well (high IoU) but struggles to declare global inconsistency because FG--BG is absent from its label space (Table~\ref{tab:hammer}). The crop–caption checker offers a light, modular signal that can be attached to HAMMER at inference-time: any splitter (e.g., SAM/YOLO or precomputed crops) feeding any captioner/encoder (we used BLIP+MiniLM) can produce an FG–BG agreement score that downstream systems may fuse with HAMMER when desired. In this paper we report the checker stand-alone to isolate its behavior; integrating it with HAMMER is deferred to future work.

\vspace{2mm}
\noindent\textbf{Takeaway.} Across heterogeneous baselines, a simple FG--BG semantic probe already surfaces global mismatch effectively and complements detectors trained on local edits.

%%%%%%%%% 6. LIMITATIONS AND FUTURE WORK
\section{Limitations and Future Work}

\textbf{Dependence on crop quality.}
The checker assumes reasonable FG/BG crops. Foregrounds leaking background (or vice versa) blur BLIP captions and inflate $\tilde{s}_{FB}$. This is a property of the upstream splitter, not of the comparator; nevertheless it affects results.

\textbf{Captioner variance.}
BLIP occasionally under-specifies scene attributes under poor lighting or unusual viewpoints. While MiniLM reduces sensitivity to phrasing, omissions or drift in crop captions remain a source of noise.

\textbf{Single-threshold decision.}
A global $\tau$ may not transfer across domains. We therefore report the continuous score $\tilde{s}_{FB}$ alongside the binary label and recommend calibrating $\tau$ on a small, domain-matched validation subset.

\textbf{Scope.}
By design, the script compares only FG vs.\ BG captions and does not consume external text or produce pixel-accurate BG masks. It is a coarse plausibility probe rather than a full detector.

\paragraph{Future work.}
We plan to (i) study robustness to different splitters (SAM, YOLO-seg, heuristics) and captioners (BLIP/OFA/CapFilt) under identical tuning; (ii) provide optional hooks for downstream fusion with HAMMER’s binary logit; (iii) explore hybrid comparators that combine text-space similarity with vision-only distances; and (iv) release small calibration sets per domain to ease threshold selection.

%%%%%%%%% 7. CONCLUSION
\section{Conclusion}

We reframed global FG--BG inconsistency detection as a modular, inference-time pipeline and instantiated its comparator \emph{exactly} as a minimal crop–caption script (BLIP-base + MiniLM), requiring only two crops per image. On an all-inconsistent split, this stand-alone probe already flags $94.3\%$ of cases with a strong F1 of $0.971$, while complementary baselines confirm that both contrastive text alignment and vision-only distances carry useful signal. As a practical extension to HAMMER, the checker provides a lightweight, pluggable scene-consistency cue that can be fused downstream without retraining, helping to close a supervision gap that arises when detectors are built solely around local manipulations.

%%%%%%%%% REFERENCES
{\small
\bibliographystyle{ieee_fullname}
\bibliography{egbib}
}

\end{document}